# Large Language Models for History, Philosophy, and Sociology of Science: Interpretive Uses, Methodological Challenges, and Critical Perspectives

Arno Simons, Michael Zichert, and Adrian Wüthrich[1]

## Abstract

This paper explores the use of large language models (LLMs) as research tools in the history, philosophy, and sociology of science (HPSS). LLMs are remarkably effective at processing unstructured text and inferring meaning from context, offering new affordances that challenge long-standing divides between computational and interpretive methods. This raises both opportunities and challenges for HPSS, which emphasizes interpretive methodologies and understands meaning as context-dependent, ambiguous, and historically situated. We argue that HPSS is uniquely positioned not only to benefit from LLMs' capabilities but also to interrogate their epistemic assumptions and infrastructural implications. To this end, we first offer a concise primer on LLM architectures and training paradigms tailored to non-technical readers. We frame LLMs not as neutral tools but as epistemic infrastructures that encode assumptions about meaning, context, and similarity, conditioned by their training data, architecture, and patterns of use. We then examine how computational techniques enhanced by LLMs, such as structuring data, detecting patterns, and modeling dynamic processes, can be applied to support interpretive research in HPSS. Our analysis compares full-context and generative models, outlines strategies for domain and task adaptation (e.g., continued pretraining, fine-tuning, and retrieval-augmented generation), and evaluates their respective strengths and limitations for interpretive inquiry in HPSS. We conclude with four lessons for integrating LLMs into HPSS: (1) model selection involves interpretive trade-offs; (2) LLM literacy is foundational; (3) HPSS must define its own benchmarks and corpora; and (4) LLMs should enhance, not replace, interpretive methods.

## 1. Introduction

Large language models (LLMs) represent a new frontier in computational research. Unlike traditional methods that rely on structured inputs, predefined categories, or handcrafted features, LLMs can flexibly process unstructured, domain-specific text. They infer meaning dynamically from context, often reducing the need to impose external interpretive frameworks or simplify complex language. Their rapid uptake across the natural and social sciences (Binz et al., 2025; Davidson and Karrell, 2025; Ziems et al., 2024) signals a broader shift in how scholarly knowledge is produced, organized, and even authored. For interpretive fields, this raises pressing questions, not only about how LLMs can support research, but also about the epistemic and cultural assumptions they embed.

---

[1] At the time of writing, all authors were affiliated with the Department of History and Philosophy of Modern Science, Technische Universität Berlin, Germany.

History, philosophy, and sociology of science (HPSS), which focuses on the historical development, conceptual foundations, and social organization of science, has long treated knowledge as context-dependent and theory-laden. As in the humanities more generally, computational methods in HPSS have often been met with skepticism, seen as trading nuance for scale or sacrificing contextual richness for abstraction (e.g., [Buchholz and Grote, 2023](); [Da, 2019]()). When *Isis* launched its 2019 Focus section on Computational History and Philosophy of Science ([Gibson et al., 2019](); [Laubichler et al., 2019]()), contributors pointed to both promise and persistent obstacles: challenges in curating structured data, the need for domain-sensitive tools, and difficulties linking statistical patterns to historical meaning. Although the "computational turn" aimed to bridge close reading and large-scale analysis, the divide remains, sustained by technical barriers and epistemic concerns (e.g., [Leydesdorff et al., 2020]()).

LLMs may mark an inflection point in the longstanding tension between computational and interpretive approaches, offering three key advances over earlier methods like topic modeling or co-word analysis. First, they provide improved semantic modeling via contextualized representations, enabling both pattern recognition and generative tasks such as paraphrasing, summarizing, or simulating discourse. Second, they are more accessible and adaptable: based on shared architectures, LLMs can be tailored to domain-specific tasks through continued pretraining, fine-tuning, or in-context learning. Combined with natural language interfaces, this lowers technical barriers for interpretive scholars. Third, LLMs bridge close and distant reading by supporting both large-scale analysis and fine-grained interpretation, making it possible to scale interpretive depth without necessarily losing contextual nuance.

For historians of science, LLMs enable large-scale tracing of conceptual change, for example, shifts in how terms like "energy", "evolution", or "objectivity" are used across scientific eras or institutions. For philosophers of science, they offer tools to map conceptual structures, analyze argumentative patterns, or simulate rival positions to test coherence, supporting work in conceptual analysis, empirically informed philosophy, or explanation. For sociologists of science, LLMs can help examine how scientific authority is constructed and contested, for instance, by analyzing how boundary-work, credibility, or dissent evolves in journals, public discourse, or policy texts. Across all three fields, LLMs not only expand methodological options but also invite renewed reflection on how meaning is inferred, knowledge structured, and epistemic claims contested. In this sense, LLMs offer a new computational affordance, one that supports rather than supplants the interpretive practices at the heart of HPSS.

In this paper, we examine the opportunities and challenges LLMs pose for HPSS research, focusing on their use as analytical tools for interpretive and empirical inquiry rather than as writing aids or general-purpose assistants. We argue that HPSS is not only well-positioned to benefit from these tools, but also uniquely equipped to critically assess and guide their use, given its reflexive stance, computational traditions, and close engagement with science and technology. Our contribution is threefold: (1) a methodological survey of current and emerging LLM applications across HPSS-relevant tasks; (2) a conceptual framework for understanding how LLMs intersect with interpretive and computational traditions; and (3) practical guidance for integrating LLMs into HPSS research workflows. In doing so, we aim to foster dialogue across HPSS and neighboring fields such as bibliometrics, science of science, computational social science, digital humanities, and computational linguistics.

This paper speaks to an interdisciplinary audience, with a primary focus on HPSS scholars. We aim to bridge technical and interpretive perspectives by pairing accessible explanations of LLM methods with critical reflection on their epistemic and methodological implications. Rather than resolving tensions between computational and humanistic approaches, we examine how LLMs can support interpretive inquiry while raising new questions about meaning, context, and infrastructure in the study of science.

The paper is structured in five parts. We begin with a brief primer on LLMs, outlining how they process language, key model types, and training paradigms. We then examine how LLMs intersect with three core methodological challenges in HPSS, following Laubichler et al. ([2019](#)): structuring data, detecting patterns, and explaining historical change. For each, we survey current practices, assess technical affordances, and reflect on how LLMs may reshape interpretive approaches. We conclude by distilling four lessons for integrating LLMs into HPSS workflows and offering a final reflection on the infrastructural and ethical conditions surrounding their development and use.

## 2. A short primer on LLMs

This section introduces how LLMs process, represent, and generate language, focusing on two core architectures: BERT-style full-context models and GPT-style generative models. We explain how token embeddings underpin model understanding, how architecture and training influence behavior, and how these choices influence both the capabilities and limitations of LLMs when working with text. Readers already familiar with transformer models may wish to skip to the HPSS-relevant applications in the next sections.

LLMs are neural networks that convert text into numerical vectors to analyze and generate language, achieving state-of-the-art performance across a wide range of natural language processing (NLP) tasks. Beyond improving familiar applications such as named entity recognition or topic modeling, they also introduce new capabilities, including fluent conversation, reasoning, and complex text-based problem-solving.

### 2.1 Embeddings and architecture

Inside an LLM, text is broken into small pieces called tokens, which include words, prefixes, suffixes, and punctuation. These are then converted into high-dimensional vectors called embeddings. These are passed through multiple layers of mathematical transformations using self-attention ([Vaswani et al., 2017](#)), where each embedding is updated based on its relation to others. The process is governed by millions of parameters or weights. These are values the model learns during training and they govern how embeddings evolve through the network. As they progress, embeddings encode increasingly rich syntactic and semantic features, becoming contextualized word embeddings (CWEs). Each layer builds a semantic space that arranges CWEs by contextual similarity, forming increasingly abstract representations across the network. Early layers capture surface features like word form or syntax; deeper layers encode more abstract, context-dependent meaning (e.g., Jawahar et al., 2019).

This approach reflects the distributional hypothesis of meaning (Harris, 1954): that meaning arises from contextual relationships. LLMs operationalize this by learning meaning as position in a high-dimensional embedding space, where semantic similarity is defined by spatial proximity.

## 2.2 Full-context vs. generative models

While LLMs are often associated with interactive dialogue, especially through chatbots like ChatGPT, not all are designed for text generation. Broadly, LLMs fall into two types: generative and full-context models, distinguished by how they process text during pretraining. This self-supervised phase trains models on large corpora to develop general linguistic competencies, conditioning internal representations and downstream capabilities. Generative models, such as GPT (Radford et al., 2018), use autoregressive training to predict each token from left to right, making each CWE part of a sequential progression. Trained to generate text word by word, these models are structurally suited for fluent, coherent output. As architectures and training data scaled up, they began producing not just well-formed language, but outputs that appeared responsive, context-sensitive, and occasionally suggestive of reasoning, emergent behaviors arising from the interaction of scale, objective, and data.

Full-context models, such as BERT (Devlin et al., 2018), rely on masked language modeling. During pretraining, random tokens are replaced with a special [MASK] token and must be inferred using words from both directions. These models apply bidirectional attention across the full input sequence, allowing CWEs to reflect a token's meaning in its wider context. This structure makes them especially effective for structured prediction tasks like named entity recognition, for generating fixed-length text embeddings used in text similarity and retrieval, and for extracting CWEs for downstream applications such as word sense modeling. Table 1 summarizes the key architectural and functional differences.

Since 2018, these models have diverged in scale and use. Generative models like GPT-4 have grown to trillions of parameters and dominate commercial applications due to their fluency and generalization. However, they are typically closed-source and resource-intensive. Full-context models, by contrast, are smaller, often open-source, and accessible for local deployment. While less flexible, they excel in structured, non-generative tasks.

|  | Full-Context LLMs (e.g., BERT, SciBERT) | Generative LLMs (e.g., GPT-4, Claude-4) |
|---|---|---|
| Architecture & Pretraining | Bidirectional (masked token prediction) | Autoregressive (left-to-right token prediction) |
| Primary Output Type | Token and text embeddings, classification scores, token-level predictions | Coherent text generation; structured outputs via natural language prompting |
| Representative HPSS Applications | Conceptual history, scientific entity extraction, citation content classification, research topic modeling | Few-shot learning of HPSS related tasks, chatting with sources (RAG), interactive "co-reasoning" |
| Accessibility | Moderate to low: requires technical setup, local compute, and ML fluency | High: accessible via web interface or API; minimal setup required |
| Transparency & Platform Dependence | More frequently open-source; models and training data often inspectable and reusable | Often proprietary and opaque; limited insight into training data or internal parameters |

**Table 1**: Key Differences Between Full-Context and Generative Large Language Models: A structured comparison of two major classes of LLMs, distinguishing their architectures, output modalities, representative HPSS applications, accessibility, and openness. These differences have direct implications for interpretive control, methodological transparency, and infrastructure demands.

## 2.3 The accessibility–literacy trade-off

Another key difference between generative and full-context LLMs lies in how they are accessed and operated. Generative models like GPT are typically designed for usability, offering intuitive interfaces that require no programming skills. This accessibility has fueled widespread experimentation across disciplines, including HPSS. A central feature enabling such flexibility is in-context learning, which allows models to perform new tasks simply by being given instructions (zero-shot learning) or examples (few-shot learning) within the prompt, without requiring fine-tuning. This process, known as prompt engineering, allows users to guide the model toward summarizing texts, extracting structured information, generating hypotheses, or classifying passages, all by crafting carefully designed inputs.

Modern generative models also increasingly support multimodal inputs (e.g., combining text with images or tables) and tool use, such as code execution, API access, or web search. These extensions, often bundled into proprietary interfaces, enable real-time access to external information sources. While powerful, they also risk giving a false sense of reliability. Fluent, confident outputs can obscure inaccuracies, omit uncertainty, or misrepresent retrieved evidence, especially in domains like HPSS where meaning is historically situated and context-dependent.

In contrast, full-context models like BERT generally require greater computational literacy to use effectively. They are typically integrated into task-specific pipelines and demand fine-tuning or post-processing to adapt to new use cases. While more transparent and open in design, their rigidity can limit exploratory analysis. BERT-style models also lack in-context learning, which reduces flexibility. Their bidirectional embeddings are useful for structured tasks such as classification or named entity recognition but are not suited to generative applications. For researchers, this presents a trade-off: more control and interpretability, but at the cost of accessibility and versatility.

Both generative and full-context models demand critical engagement: while generative models offer accessibility, they risk obscuring issues of reliability and domain fit; full-context models provide more control but require technical expertise. For HPSS researchers, the challenge lies in balancing usability with epistemic rigor.

## 3. Data, models, and training

In LLM-based research workflows, the role of data shifts in subtle but significant ways. Rather than requiring fully structured inputs from the outset, these models are designed to extract and model meaning from unstructured text. As a result, interpretive choices increasingly take place within model design, training, and prompting, reframing traditional computational challenges of data curation as questions of model adaptation and interpretive fit.

### 3.1 Data complexity and model assumptions

Advocating the computational turn in HPSS just before the rise of LLMs, Laubichler et al. ([2019](#)) identified the provision and curation of structured data as a central challenge. HPSS data, they argued, are often fragmented, inconsistently formatted, and hard to structure without sacrificing nuance. More than a technical issue, this is an epistemological one: knowledge is historically situated, and so are the categories, concepts, and evidentiary standards embedded in data. What counts as meaningful shifts over time, conditioned by changing scientific practices and cultural contexts. Data curation must therefore contend with both the fragmented form and historical content of knowledge.

LLMs appear to shift some of these challenges by working well with unstructured text, processing raw language without predefined categories or formats. But this flexibility brings new complexities. The model itself becomes an epistemic infrastructure: not just a tool trained on data, but a condensed representation of large text corpora, shaped by decisions about inclusion and encoding, and embedded in the workflows, assumptions, and institutions that structure knowledge production. This raises concerns for HPSS, where much of the material comes from earlier periods. LLMs trained mainly on contemporary data may flatten or misrepresent historically specific language and concepts, particularly in archival texts, outdated vocabularies, or shifting conceptual frameworks.

As a result, we must attend not only to data, but also to how models are trained and what assumptions they encode. At the same time, structured data remains crucial, not just for fine-tuning or grounding LLMs via sources like knowledge graphs or citation networks, but also as a primary object of inquiry. HPSS scholars will continue to build and analyze structured datasets to trace conceptual change, map intellectual networks, and make historical claims beyond what LLMs can currently infer. While LLMs may assist in curating such data from unstructured text, the interpretive work of designing and analyzing them remains central. The data challenge hasn't disappeared—it has changed form.

The materials HPSS researchers engage with span a wide spectrum, from technical scientific articles to science-policy documents, media discourse, field notes, and multimodal artifacts like figures and diagrams. Each genre brings its own complexities, and many sources also pose temporal challenges, with historical or shifting language and conceptual vocabularies. No single model or system can meet all these demands, and careful model selection and adaptation are essential.

## 3.2 Domain-specific pretraining

A key strategy for adapting LLMs to specialized domains is domain-specific pretraining, where models are exposed to targeted corpora during their initial learning phase. This structures internal representations and constrains the embedding space in ways that persist, influencing how models interpret and generate language. While full pretraining is resource-intensive, a common alternative is continued pretraining: further training an existing model like BERT or LLaMA-2 (Touvron et al., 2023) on domain-specific texts that were underrepresented in the original corpus. Pioneered by models like BioBERT (Lee et al., 2020) and SciBERT (Beltagy et al., 2019), this approach has also shown recent promise for HPSS-specific applications (Simons, 2024; Zichert et al., 2025).

Pretraining a model from scratch on scientific data was first explored with a variant of SciBERT and later adopted by models like PubMedBERT (Gu et al., 2021), BioGPT (Luo et al., 2022a), and Galactica (Taylor et al., 2022). This approach offers greater adaptability by avoiding biases from general-purpose models. Unlike continued pretraining, which retains the base model's vocabulary, from-scratch pretraining enables vocabulary customization to better represent specialized terms. It also allows architectural changes, such as adapting attention mechanisms for temporal reasoning (Rosin and Radinsky, 2022) or building multimodal models from the ground up (Wang et al., 2023c). However, given its high data and computational demands, from-scratch pretraining remains impractical for most HPSS applications. For a comprehensive overview of targeted pretraining for scientific texts, see Ho et al. (2024) and Zhang et al. (2024b).

## 3.3 Task-specific fine-tuning

Beyond pretraining, LLMs can be adapted to HPSS-specific purposes through fine-tuning on particular NLP tasks. In NLP, a "task" refers to a defined goal, such as sentence classification, question answering, or named entity recognition, where the model must act on text in specific ways. Fine-tuning for such tasks typically involves supervised learning on datasets that pair inputs with expected outputs. This requires carefully labeled examples, often informed by human labor and interpretive judgment. In HPSS contexts, where meaning is historically situated and categories are fluid, the assumptions built into labeled data carry particular weight.

Three widely used strategies include adding task-specific classification layers, applying prompt-based fine-tuning, and using contrastive learning, as exemplified by science-specialized models such as BioBERT, BioGPT, and SPECTER (Cohan et al., 2020), respectively. Table 2 outlines how these approaches fit into the wider training landscape and highlights potential HPSS applications.

Task-specific classification typically adds one or more layers to a pretrained model to map token or span representations to predefined labels. BioBERT uses this approach for biomedical tasks such as named entity recognition and relation extraction, learning to identify patterns like gene–disease associations in annotated data. These added layers transform CWEs into task-specific predictions, guided by labeled training examples.

Prompt-based fine-tuning treats structured tasks as text generation problems, using natural language prompts instead of added output layers (Liu et al., 2023). The task is encoded directly in the input, typically by researchers using templates. For example, BioGPT was trained on pairs like: Input: "What is the relationship between aspirin and COX-1?"; Output: "Aspirin inhibits COX-1".

Contrastive learning fine-tunes models on text pairs labeled as similar or dissimilar to produce fixed-length text embeddings that reflect semantic similarity, supporting tasks like document clustering or retrieval. SPECTER, for example, builds on SciBERT and was trained using citation links as a proxy for similarity, avoiding manual labeling, but inheriting whatever kinds of similarity citations actually encode. Text embeddings from models like SPECTER or more advanced Sentence-Transformers ([Reimers and Gurevych, 2019](#)) now underpin many methods entering HPSS, including thematic clustering and novelty detection.

By relying on labeled training data, these strategies encode specific assumptions about meaning, relevance, and relation, which carry particular weight in HPSS contexts. When applying tools like BioBERT, BioGPT, or SPECTER to historical inquiry, researchers must account for the temporal mismatch between the models' training data and their sources. And as discussed next, common similarity metrics driving text embeddings may diverge from the interpretive concerns of a given HPSS investigation.

## 3.4 Text embeddings: Design choices and interpretive consequences

Text embeddings, such as sentence- or document-level vectors, are widely used in LLM pipelines for text comparison or retrieval. But they are not neutral representations of meaning. Instead, they are trained to reflect particular notions of similarity, defined, for instance, through paraphrase relationships, shared citations, topical overlap, or domain proximity. Each of these training setups relies on specific assumptions about what makes two texts alike. These assumptions, often implicit, influence the structure of the embedding space and carry interpretive consequences for downstream use, especially in historical or cross-contextual settings where notions of similarity are themselves variable and contested.

A useful contrast lies in how token and text embeddings are created. Token embeddings are learned from large corpora by modeling how words tend to co-occur. Based on billions of distributed language choices, these patterns are emergent and largely unintentional, driven by general use rather than explicit meaning. Text embeddings, by contrast, are fine-tuned on smaller, more focused datasets, often labeled by annotators or structured by patterns like citations. These involve fewer, more deliberate choices aimed at capturing specific types of similarity, such as topical overlap or scientific relatedness. This offers model designers more control but narrows the range of relationships the model can represent. This added focus can improve performance on certain tasks or alignment with domain-specific goals, but also risks oversimplifying ambiguity, flattening conceptual nuance, and encoding assumptions that remain opaque to users.

For HPSS scholars, this has direct methodological implications. Rather than merely discovering relationships, text embedding models construct them, guided by technical choices about similarity, relevance, and relatedness. For example, when building scientific novelty detectors based on dissimilar abstracts (see [section 5.2](#)) citation-based abstract embeddings may be problematic: while citations can signal intellectual connection, they also reflect institutional ties, disciplinary norms, or rhetorical strategies ([Bornmann and Daniel, 2008](#)). If such embeddings capture conceptual divergence at all, they do so in highly mediated ways that are difficult to disentangle. More generally, whenever text embeddings are used they carry interpretive assumptions that structure what becomes visible and what remains hidden.

|  | Core Idea | Data Required | Representative HPSS Application |
|---|---|---|---|
| Domain-Specific Pretraining | Exposing models to HPSS-specific language during pretraining (from scratch or continued) | Large domain-specific corpus | Capturing field- or time-specific vocabularies and semantics in scientific texts |
| Task-Specific Fine-Tuning | Training a model on supervised data for classification, NER, etc. | Labeled examples per task | Scientific entity and relation extraction, citation context classification, scientific argument mining |
| Contrastive Fine-Tuning | Optimizing pooled embeddings by training on pairs of similar and dissimilar sequences | Similar/dissimilar sentence pairs or documents | Research topic modeling, novelty detection, revision tracking |
| Prompt-Based Learning | Framing tasks as instructions or examples in prompts (no weight updates) | None or a few illustrative examples (<100) | Few-shot learning of HPSS related tasks, interactive "co-reasoning" |
| Retrieval-Augmented Generation (RAG) | Augmenting generation with external document retrieval | External corpora + retrieval embeddings | Chatting with sources, research topic summarization |

**Table 2:** Strategies for Adapting Large Language Models to HPSS Research Contexts. A comparative overview of five major adaptation strategies: domain-specific pretraining, task-specific fine-tuning, contrastive learning, prompt-based learning, and retrieval-augmented generation, focusing on their methodological principles, data requirements, and representative applications in HPSS. The table emphasizes the interpretive trade-offs and infrastructural implications associated with each approach.

## 3.5 Retrieval-augmented generation (RAG) and tool use

Beyond pretraining and fine-tuning, retrieval-augmented generation (RAG) and related tool-based approaches offer flexible ways to adapt LLMs to the needs of HPSS research. In a typical RAG setup, a generative model is paired with an external retrieval system, often using a BERT-style similarity model, to fetch relevant texts that the LLM then incorporates into its output (Lewis et al., 2020). This allows the model to access external knowledge dynamically, rather than relying solely on pretrained information.

For instance, Med-PaLM 2 (Singhal et al., 2025) retrieves current medical literature to support clinical question answering, helping ground its outputs in up-to-date science. GeoGalactica (Lin et al., 2024) uses tool learning to access geoscience databases and computational tools, refining answers with real-time data.

RAG features are also appearing in scholarly platforms. Dimensions' Research GPT (via ChatGPT) delivers context-aware responses with clickable citations from publications, clinical trials, and grants. Web of Science's Research Assistant uses natural language queries and a knowledge graph to reveal conceptual links across multilingual sources. Scopus is rolling out tools like Semantic Reader to summarize and highlight key insights in journal articles.

As such systems become more widespread, both in domain-specific tools and general platforms like ChatGPT, Gemini, or Claude, they offer HPSS scholars new ways to engage with sources through natural language. Moving beyond keyword search, these tools operate semantically, enabling more intuitive retrieval and dynamic interaction: scholars can pose questions, explore meanings, and contextualize claims dialogically. RAG-powered systems may thus serve as interpretive partners, enabling researchers to "chat with papers" (Tykhonov et al., 2025), exploring structure, content, and conceptual implications in iterative ways.

These affordances, however, introduce epistemic risks. With more functionality embedded in natural language interfaces, interpretive assumptions are often obscured, embedded within retrieval algorithms, similarity metrics, or prompt templates. This opacity may conflict with HPSS values, reinforcing the need for transparency and critical scrutiny in how such systems are adopted and used.

As this overview shows, the challenges identified by Laubichler et al. (2019) have not disappeared—they have shifted. LLMs reduce the need for highly structured data, but introduce new issues around transparency, reproducibility, and infrastructure control. Proprietary models often obscure their training data and design choices, limiting scrutiny. Open-source alternatives are more transparent but require significant computational resources, unevenly distributed across institutions. In this landscape, LLM literacy becomes essential: not just technical skill, but interpretive awareness of how models work, what assumptions they encode, and how they shape inquiry. These issues also raise broader questions of accountability and the politics of research infrastructure. Meanwhile, emerging multimodal models, able to process text alongside images, diagrams, and other formats, open a new frontier (Wang et al., 2023c). These may prove especially valuable for HPSS, where scientific texts often interweave with visual, material, and spatial representations.

## 4. Patterns

This section addresses Laubichler et al.'s (2019) second key methodological challenge: how to detect and interpret patterns in scientific knowledge at scales beyond traditional close reading. While deep learning was already seen as promising at the time, the rise of LLMs marks a significant advance in this trajectory. These models can identify patterns that are not only larger in scale, but also more contextually embedded and semantically rich. In HPSS, such patterns include both formal and informal structures that organize scientific knowledge: conceptual taxonomies, disciplinary classifications, thematic groupings, and genre-specific features such as section headings, citation practices, metadata, and authorship conventions.

LLMs support the analysis of such patterns through tasks like topic modeling, named entity recognition, and relation extraction, often used in overlapping or hybrid workflows. As with earlier models like word2vec, the outputs of LLMs are mediated not only by the input text, but also by the model's learned representation and training data. In BERT-like models, this mediation appears in how texts are embedded or classified; in GPT-like models, it extends to generative outputs shaped by prompts and predictive sequencing. While both can flatten distinctions or reinforce spurious associations, generative models may also hallucinate, producing fluent but fabricated content that reflects training priors more than actual sources. These dynamics introduce new challenges for interpretive work.

Recent applications show how LLMs are being used to identify and model HPSS-relevant patterns such as scientific concepts, research themes, and citation roles. These efforts vary in approach and linguistic scale, from individual words to entire documents. Broadly, they fall into two categories: exploratory uses that surface latent patterns without predefined targets, and targeted uses that detect specific categories or features. These roughly correspond to unsupervised and supervised methods in NLP, respectively, and span tasks involving token-level semantics, sentence or paragraph analysis, and multi-token structures such as hedge phrases or causal links. The subsections that follow are organized around this combined framework.

## 4.1 Exploration of unknown patterns

As noted above, a core innovation of LLMs is their ability to represent tokens (usually words or subwords) in context as numerical vectors. These CWEs form the foundation for everything LLMs do, encoding not only syntax but also the context-specific meaning of words. This makes them particularly valuable for HPSS research concerned with ambiguity, conceptual variation, and the situated use of language. At the token level, the primary exploratory use of CWEs has been in word sense modeling. This typically involves clustering to identify distinct senses of polysemous terms, either by directly clustering CWEs of word occurrences (Wiedemann et al., 2019), or by clustering word substitutes derived from CWEs via masked language modeling (Amrami and Goldberg, 2019).

Despite their potential, such methods remain underused in HPSS contexts. Kleymann et al. (2022) applied CWE clustering to the word "theory" across nearly 4,000 digital humanities articles. Although the study aimed to identify sense clusters, the authors found that the results primarily captured syntactic variation rather than meaningful semantic distinctions. Simons (2024) analyzed clusters of "Planck" embeddings using both general and domain-adapted BERT models on 100,000 physics-related Wikipedia articles and 1,500 academic papers. In contrast to Kleymann et al.'s work, these clusters matched expected senses well and revealed that domain-specific models better captured fine-grained distinctions. To study conceptual variation in scholarly jargon, Lucy et al. (2023) used a substitute-based word sense induction method. ScholarBERT predicted the top five substitutes for 4,000 lemmatized target words, which were used to build co-occurrence graphs and clustered into word senses via Louvain community detection. They identified clear discipline-specific senses for many terms, such as "bias" in statistics, psychology, and climate science, but noted challenges with clustering granularity and high computational demands.

For historians of science, these techniques could, in principle, help identify divergent uses of key terms across corpora, helping map conceptual variation across disciplines or schools of thought. Philosophers may use them to analyze shifts or ambiguities in the usage of foundational concepts within theoretical debates. For sociologists, these methods may offer a way to detect terminological distinctions that may correlate with institutional affiliations or disciplinary boundaries.

Moving beyond token-level analysis, the core strategy for exploring semantic patterns in unstructured text involves clustering fixed-length text embeddings to identify latent topical structure. This approach uses sentence- or document-level embeddings generated by models such as Sentence-BERT or SPECTER and applies clustering algorithms to uncover emergent groupings based on semantic similarity. Tools like BERTopic (Grootendorst, 2022), which combine these embeddings with dimensionality reduction and topic ranking, have become especially popular due to their flexibility and accessible design. BERTopic has emerged as a serious alternative to existing topic modeling technologies, offering comparable or superior topic coherence in many settings. In HPSS, it has been applied to map thematic structures both in scientific corpora (e.g., Kim et al., 2024) and public discourse (e.g., Falkenberg et al., 2022), enabling exploratory analysis of conceptual variation across texts. We return to further HPSS-specific applications in the next section, where BERTopic's use in modeling discursive change over time is examined in more detail.

While BERTopic currently dominates LLM-based topic modeling in HPSS contexts, several alternatives have been explored. These include hybrid models that combine BERT embeddings with traditional approaches like LDA or variational autoencoders (Bianchi et al., 2021; George and Sumathy, 2023), as well as FASTopic (Wu et al., 2024), which links documents, topics, and words semantically. BERTopic performs comparably to, or better than, VAE-based models in terms of topic coherence, while being simpler and more efficient (Grootendorst, 2022; Zhang et al., 2022). Yet its assumption of a single dominant topic per document can obscure conceptual multiplicity (Egger and Yu, 2022), and its results are sensitive to clustering parameters and dimensionality reduction. As with all embedding-based methods, it also inherits biases from general-purpose training corpora, which may misrepresent historical language or domain-specific concepts. FASTopic promises richer topic distributions and improved interpretability, while newer generative models like TopicGPT (Pham et al., 2023) and PromptTopic (Wang et al., 2023d) allow more flexible, human-readable output, though they come with higher computational cost. Looking ahead, hybrid frameworks that combine embeddings, prompting, and probabilistic modeling may hold promise for HPSS applications

## 4.2 Detection of known patterns

In contrast to exploratory approaches, which let patterns emerge from data, many HPSS-relevant tasks involve the detection of predefined categories or structures. These range from token-level classification (e.g., word sense disambiguation), to span-level detection (e.g., citation context extraction), to document-level classification (e.g., rhetorical function or disciplinary field). Combined appropriately, these techniques can recover complex structures such as entity networks, argumentative sequences, or causal relations. To perform these tasks, researchers typically fine-tune pretrained LLMs on labeled datasets, especially full-context models like SciBERT. Alternatively, generative models such as GPT-4o can be prompted directly in few-shot or zero-shot settings, enabling task performance without extensive retraining.

Citation context analysis is a prominent use case, especially within bibliometrics. Here, LLMs are used to classify the rhetorical intent (e.g., background, method, critique) or sentiment of citations (e.g., agreement, disagreement), building on benchmark datasets like SciCite, ACL-ARC, and 3C (Roman et al., 2021; Jiang and Chen, 2023). Large-scale platforms like scite (Nicholson et al., 2021) have operationalized this task in production settings, using proprietary taxonomies to classify millions of citation contexts. While fine-tuned models dominate this space, recent work explores prompting as a lightweight alternative ([Kunnath et al., 2023](#)).

Citation context extraction, which involves identifying the parts of a document that explain a citation's role, is closely linked. It has evolved from early fixed-window methods that often included irrelevant text or missed key context. Kunnath et al. ([2022](#)) addressed this with a dynamic approach using transformer-based embeddings (e.g., SPECTER, SciNCL) to select context based on semantic similarity, improving classification in heterogeneous corpora. The FOCAL shared task ([Grezes et al., 2023](#)) extended this by requiring models to both classify citation function and extract the supporting span, showing that robust citation analysis hinges as much on context selection as on model design.

Current approaches are often driven by science-policy goals, focusing on tasks like quality assessment or citation recommendation, rather than supporting critical, reflexive analysis. This is particularly evident in how the labels and examples for training are defined, often with limited attention to the contextual or contested nature of citation. For HPSS purposes, we suggest that these tools would benefit from closer engagement with interpretive traditions that understand citation as a social and epistemic practice ([Gilbert, 1977](#); [Latour, 1987](#))

Beyond citation analysis, LLMs have been widely used in scientific domains to extract information like domain-specific entities or material–property associations ([Dagdelen et al., 2024](#); [Ji et al., 2000](#)). They have also been applied to map argumentative and causal dependencies ([Fergadis et al., 2021](#); [Gorour et al., 2024](#); [Zhang et al., 2024b](#)), and to classify documents by topic, contribution type, or structural role ([Chen et al., 2022a](#); [Chen et al., 2022b](#); [Ma et al., 2022](#)). Across these applications, fine-tuned domain-specific models like BioBERT, SciBERT, and MatBERT consistently outperform general-purpose baselines on such tasks ([Beltagy et al., 2019](#); [Dagdelen et al. 2024](#); [Ji et al., 2000](#); [Lee et al., 2020](#); [Luo et al., 2022a](#)). GPT-style models, in contrast, perform well in zero- and few-shot scenarios when given clear prompts. Their advantages lie in accessibility and adaptability; their weaknesses include domain specificity, factual unreliability, contextual ambiguity, high computational costs, and integration complexity ([Shao et al., 2024](#); [Zhu et al., 2024b](#)). Thinking this further, there is clear potential to apply these methods, or combinations of them, to extract structured data from unstructured HPSS corpora ([Dagdelen et al. 2024](#), [Zhu et al. 2024a](#)), addressing one of the core challenges highlighted by Laubichler et al. ([2019](#)).

The LLM-based techniques for pattern detection discussed in this section offer powerful tools for surfacing conceptual, rhetorical, and thematic structures across scientific texts. In HPSS research, they can help trace shifts in discourse, map argumentative patterns, and extract entities such as authors, instruments, or concepts, which can then be integrated into historical or sociological analyses. These methods rely on the distributional premise that meaning arises from context-sensitive relationships among words, a view that resonates with interpretive traditions but also risks oversimplification when taken uncritically. In structured tasks with limited ambiguity, LLMs may support relatively robust, automatable insights; in more open-ended interpretive contexts, they can surface plausible framings or discursive patterns that warrant further analysis. Their value lies not in offering definitive classifications, but in expanding the range and scale of what can be noticed, compared, and questioned. They enrich interpretive judgment through computational means while remaining attentive to the contextual and contested nature of meaning in scientific discourse.

## 5. Dynamics

The third major methodological challenge identified by Laubichler et al. ([2019](#)) is explaining scientific change. While the earlier challenges—data structuring and pattern detection—concern how knowledge is represented and recognized, this one asks how it evolves: how concepts, practices, and institutions emerge, shift, or dissolve over time. For HPSS scholars, such questions have traditionally been addressed through contextual analysis and critical source work. This section asks whether and how LLMs might contribute to that task.

We distinguish two complementary levels at which LLMs can model scientific change: token-level dynamics, capturing shifts in the meaning or use of individual terms, and text-level dynamics, modeling broader discursive patterns such as topic emergence, argumentative change, or evolving citation practices. At both levels, LLMs offer new ways to access diachronic patterns, surfacing trends that might escape close reading. But as with other challenges identified by Laubichler et al., these affordances come with epistemic costs: issues of interpretability, operationalization, and historical alignment persist. What counts as semantic change? When does a textual shift reflect a conceptual rupture? And can statistical similarity serve as a proxy for historical continuity?

The sections that follow survey LLM-based approaches to both levels, examine their assumptions, and assess their relevance for interpretive HPSS research.

### 5.1 Token-level dynamics

At the most granular level, LLMs can model scientific change through token-level analysis, tracing how individual words or concepts shift in meaning, usage, or connotation over time. By comparing CWEs across temporally segmented corpora, researchers can study lexical semantic change (LSC) and the evolution of disciplinary vocabularies ([Periti and Montanelli, 2024](#)). These models build on earlier co-occurrence and distributional methods ([Gavin et al., 2016](#); [Wevers and Koolen, 2020](#)), but offer finer-grained, context-sensitive representations of meaning.

Recent HPSS applications reveal both the potential and limits of these methods. Kleymann et al. ([2022](#)) tracked shifting semantic similarity between "theory" and related terms like "model" and "method" in digital humanities over five decades, using averaged CWEs to measure semantic drift. Zichert et al. ([2025](#)) analyzed the evolving use of "virtual" as a linguistic marker of the virtual particle concept in physics, applying a domain-adapted BERT model trained on nearly a century of *Physical Review* publications. By combining time-sliced embeddings with clustering, distance metrics, and entropy-based measures, they assessed both shifts in dominant meaning and growing semantic variability, i.e., polysemy, over time. Simons ([2024](#)) modeled shifts of "Planck" in arXiv physics papers over 30 years, combining similarity metrics with interpretive analysis. These studies show how LLMs can support historically grounded concept tracing, especially when paired with close reading or expert validation.

Still, several challenges remain. Embedding shifts can reflect contextual noise rather than true semantic drift ([Kutuzov et al., 2022](#)), and efforts to classify change types ([Cassotti et al., 2024](#); [Periti and Montanelli, 2024](#)) or assess statistical significance ([Liu et al., 2021](#)) are still developing. Temporal modeling adds complexity: single models may blur distinctions over time, while separately trained models pose alignment issues. Alternatives like embedding temporal markers or tracking sense clusters remain underused but promising ([Periti and Montanelli, 2024](#)).

These technical issues also reflect deeper conceptual tensions. What counts as a meaningful shift in usage? Current BERT-based approaches typically focus on individual terms and their local context, whereas conceptual history often concerns semantic fields or evolving networks of meaning ([Gavin et al., 2016](#); [Wevers and Koolen, 2020](#)). This mismatch complicates interpretation and highlights the ongoing need for humanistic judgment.

Evaluating LSC models remains challenging particularly in multilingual settings (e.g., early quantum mechanics in both German and English), in the face of limitations and biases in historical corpora, and when dealing with genre-specific features such as formulas or non-standard syntax. Existing benchmarks are largely designed for general-purpose tasks and rarely capture the ambiguity or domain-specificity central to HPSS research. As a result, qualitative validation remains essential: in studies like Simons ([2024](#)) and Zichert et al. ([2025](#)), interpretive reading was key to assessing whether observed shifts reflected meaningful conceptual developments.

Generative models like ChatGPT have recently been explored for semantic shift detection ([Periti et al., 2024](#)), but their outputs remain inconsistent and less reliable for fine-grained diachronic work. For now, BERT-style models offer better control and interpretability, although hybrid methods that combine embeddings with temporally enriched or network-based models may hold particular promise for future HPSS research.

Despite these challenges, token-level semantic modeling remains a promising tool for HPSS inquiry. Historians might use it to trace conceptual histories within or across scientific disciplines; philosophers could explore how meaning variation relates to conceptual ambiguity or theory change; sociologists may examine how the use of key terms indexes transformations in institutional discourse, disciplinary boundaries, or professional identity.

## 5.2 Text-level dynamics

Beyond individual tokens, many HPSS questions focus on how larger discursive structures such as the emergence of new ideas, shifts in argumentation, and transformations in scholarly communication develop and circulate over time. Capturing these dynamics requires modeling strategies that operate across multiple textual scales: from sentences and paragraphs to full arguments, documents, and corpora. In what follows, we examine LLM-based approaches to these phenomena across three interrelated areas: dynamic topic modeling, scientific novelty detection, and the analysis of citation, influence, and revision. Each addresses a distinct facet of textual evolution, from the formation of shared vocabularies to the transformation of ideas.

Methods in these areas draw on a range of modeling strategies, including full-context architectures (e.g., BERT-based embeddings combined with clustering or similarity metrics), generative approaches (e.g., in-context prompting), and hybrid pipelines that combine both. Regardless of technique, these models rely, implicitly or explicitly, on internal representations of textual similarity and difference. Even when not computing distances directly, generative models draw on high-dimensional relational structures learned during training, shaping how they simulate continuity, rupture, or emergence in discourse.

Dynamic topic modeling traces how research themes evolve by clustering semantically similar texts within temporally segmented corpora. Recent approaches often rely on LLM-derived embeddings, such as BERT or SciBERT, to generate topic structures that reflect contextual meaning rather than surface co-occurrence. Among these, BERTopic has become a popular tool for modeling discursive change, thanks to its flexibility, built-in dynamic capabilities, and integration with modern embedding pipelines. For example, Wang et al. (2023b) used BERTopic to map interdisciplinary topic trajectories in library and information science, combining topic evolution with diversity and cohesion metrics, while Wang et al. (2023a) applied dynamic BERTopic modeling to analyze narratives around AI in international newspapers over 12 years.

Scientific novelty detection aims to identify contributions that diverge meaningfully from prior discourse (Zhao and Zhang, 2025). One approach models novelty as semantic deviation, using embedding-based similarity to flag outliers. For example, Luo et al. (2022b) measured distances between embeddings of research questions and methods, while Just et al. (2024) used cosine similarity across documents. Another treats novelty as a learnable feature, using supervised models trained on citation cues or labeled data. Song et al. (2023), for instance, combined BERT embeddings with patent structures to detect emerging technology clusters. Recent work explores generative methods such as prompt-based scoring (de Winter, 2024; Bornmann et al., 2024) or linguistic surprise measures (Vicinanza et al., 2023). Across these approaches, novelty is treated as a signal in semantic space. Yet for HPSS, this invites caution: epistemic innovation is often incremental, contested, or reframed. LLM-based novelty detection may surface useful candidates, but their significance must be interpreted historically and conceptually, not inferred from distance metrics alone.

Dynamic citation analysis, influence modeling, and revision tracking seek to trace how ideas propagate, shift, and interact across documents through citation, paraphrase, argumentation, or editorial change. Arnaout et al. (2025) used prompt-based learning to classify "impact-revealing" citations and summarize how a paper's reception evolves across the citing literature. Cheng et al. (2024) linked linguistic similarity to citation lag using article embeddings. Lin et al. (2020) applied BERT-based classification to compare preprints with final publications, showing that more extensive revisions, especially to abstracts and introductions, correlate with eventual acceptance. Gorur et al. (2024) tested generative LLMs on support/attack classification in non-scientific texts, finding strong few-shot performance but challenges with subtle disagreement. Li (2024) used sentence embeddings to trace influence via paraphrased references in historical texts. Jiang et al. (2022) used full-context classifiers to align and compare arXiv versions and categorize revision types. Collectively, they demonstrate how LLMs can illuminate evolving scientific discourses at scale, revealing not just who cites whom but how ideas are reframed, contested, and transformed across time and context

The methods surveyed here, from modeling semantic shift to analyzing novelty, influence, and revision, remain largely prospective for HPSS, having emerged mainly from bibliometrics and NLP. Like other forms of modeling, they rely on abstraction, idealization, and representational choices (e.g., Jacquart et al., 2023; Knuuttila, 2011), compressing linguistic data into semantic spaces influenced by assumptions about continuity and relevance. While these approaches offer powerful tools for studying scientific change at scale, their training setups may fail to resonate with the ambiguity and historical specificity central to HPSS.

Looking ahead, LLMs may also enable new forms of agent-based modeling in HPSS, simulating communicative, context-sensitive actors in dynamic environments (Guo et al., 2024). This echoes Laubichler et al.'s (2019) call to model scientific change via formal representations of agent interaction. While emerging models promise narrative-rich simulations, it remains unclear whether they can capture the situated reasoning and historical contingency central to HPSS. Rather than resolving epistemic challenges, they may repackage them in more opaque, computation-heavy forms (Larooij and Törnberg, 2025). These possibilities are intriguing, but call for careful scrutiny.

## 6. Discussion

Before turning to our four key lessons, we offer two brief examples of how the methods above might combine into concrete HPSS workflows, showing how LLMs can help explore longstanding questions. We focus on two: how epistemic change unfolds over time, and how knowledge moves across disciplinary or institutional boundaries. Traditionally addressed qualitatively, both can now be studied through scalable pattern-tracing across large corpora.

To examine epistemic change, one might combine topic modeling to track research themes with CWEs to trace shifts in meaning over time. Sudden topic shifts or semantic drift could indicate conceptual breaks, while gradual changes may suggest incremental development. This would build on work by Simons (2024), Zichert et al. (2025), and Wang et al. (2023a), who use embeddings and topic models to study conceptual change.

To study knowledge transfer, one might extract cross-disciplinary citations, classify their rhetorical roles, and compare citing and cited language. This could reveal how concepts are translated, reframed, or strategically redeployed across fields. Citation function analysis (Kunnath et al., 2022; Arnaout et al., 2025; Leto et al., 2024) and semantic similarity modeling (Luo et al., 2022b; Li, 2024) offer tools for tracing how epistemic resources shift meaning in context.

Across both examples, scientific entity extraction and knowledge graph construction (Dagdelen et al., 2024; Ji et al., 2000) could help map relationships among concepts, actors, and methods. RAG approaches, including "chatting with papers" systems (Tykhonov et al., 2025), could support iterative, source-grounded exploration of term use, argument evolution, and conceptual shifts.

Unlike earlier computational methods, LLMs enable richer, more flexible forms of modeling and interpretation. They offer a way to realize the promise of the computational turn: not as a turn away from interpretation, but toward its systematic extension. Rather than displacing interpretive work, LLMs may help scale it, scaffold it, and reflect on its assumptions, bringing new methodological depth to core HPSS questions.

With that in mind, we now turn to four lessons to guide the critical and constructive integration of LLMs into HPSS research.

## 6.1 Lesson 1: Model selection comes with technical and interpretive trade-offs

Model choice in LLM-based research is never neutral. It entails trade-offs between technical feasibility and interpretive fidelity. For HPSS scholars, this means balancing practical factors like access, performance, and transparency with the epistemic and methodological goals of the project.

Full-context models like BERT and its variants (e.g., SciBERT) are efficient and adaptable, particularly for structured tasks like classification or retrieval. They support fine-tuning and often come with open-source benefits, but require technical expertise and labeled data. Generative models like GPT-4 or Claude-4 offer more flexibility and accessibility, especially for exploratory tasks. Yet their proprietary nature, tendency to hallucinate, and opaque logic pose challenges for reproducibility and interpretability.

These technical distinctions carry interpretive consequences. Some models generalize well but flatten domain-specific distinctions. Others may perform accurately but lack the semantic flexibility to capture conceptual ambiguity or diachronic change, which are central concerns in HPSS. For instance, a general-purpose generative model may fluently summarize a 19th-century scientific text, but misrepresent its conceptual vocabulary due to unfamiliarity with historical context. Likewise, a fine-tuned BERT variant may accurately extract scientific entities, while being of no use for understanding the rhetorical or strategic function those entities serve in justifying a method or conclusion.

In many cases, hybrid strategies will likely prove most effective, for example, using a full-context model to detect patterns in the usage of key terms across time, and then a generative model to interpret those shifts by examining how the terms are framed in contextually significant passages. The key is not tool loyalty, but awareness of what each model assumes and enables, and how their interaction influences the research.

Prospectively, multimodal models could expand this toolkit further, especially for analyzing visual elements in scientific communication. As always, these affordances must be critically assessed in light of the interpretive values that define HPSS.

## 6.2 Lesson 2: LLM literacy is foundational for interpretive research

As LLMs become more integrated into research, interpretive scholars increasingly encounter their outputs. Developing a degree of LLM literacy is becoming essential for anyone engaging with LLM-assisted research, as it involves understanding how these models process language, generate responses, and embed assumptions.

This literacy operates on multiple levels. Conceptually, scholars should understand the basics of transformer-based models: how they tokenize language, compute contextual embeddings, and generate meaning via self-attention. This includes distinctions between full-context models (e.g., BERT) and generative ones (e.g., GPT), and how models are steered through fine-tuning, prompt engineering, or RAG. Practically, it means recognizing how model behavior reflects prompt design, training data, and architectural constraints. Even scholars not using LLMs directly should be able to interpret outputs critically and see how technical choices condition interpretive claims. For LLM users, literacy also involves hands-on familiarity with model interfaces and their role in workflows like classification or clustering. The goal is not technical mastery, but informed engagement, i.e. knowing how models filter and frame both questions and answers.

LLM literacy extends the practices of critical reading. These models do not simply reflect language use. They also operationalize specific assumptions about meaning, context, and similarity, structured by training data and architectural design. As emerging epistemic infrastructures, they increasingly influence how language is processed and knowledge is represented. Just as scholars interrogate the framing of a historical source or the argumentative structure of a philosophical text, so too must they examine model outputs with interpretive care. What is being measured? Why is a concept clustered or linked in a particular way? What kinds of distinctions or ambiguities are elided? Attending to these questions allows researchers to treat model outputs not as self-evident truths, but as situated interpretations, contingent on the model's construction and the epistemic frameworks it encodes.

Without such literacy, there is a risk of epistemic outsourcing, treating fluent outputs as authoritative and technical pipelines as inherently objective. With it, researchers are better positioned to engage LLMs critically and creatively, not to replace expertise but to extend it. Cultivating this literacy, whether through direct training or collaboration, is therefore essential.

## 6.3 Lesson 3: HPSS must define its own benchmarks and corpora

To make LLMs useful for HPSS, scholars must help develop the datasets and evaluation frameworks that guide their development. Most existing models are trained on general or technical domains and assume fixed labels and stable taxonomies, assumptions that don't hold in HPSS, where sources are historical, vocabularies shift, and ambiguity is often central.

Ambiguity in HPSS arises both from the content of scientific texts and from the frameworks scholars use to interpret them. Operationalizing concepts like "research topic" or "conceptual change" often exposes disagreement, pluralism, or fluidity that resist standard annotation. Some of this can be managed through richer annotation strategies (e.g., multi-labeling, confidence scoring), but much of it requires models that support interpretation rather than enforce consensus. This affects workflow design: while supervised methods rely on labeled data, unsupervised methods like topic modeling or semantic shift detection can embrace ambiguity more productively, provided their limits are understood.

In HPSS, evaluation is not merely a technical step but an interpretive act. Standard metrics may suffice for benchmarking some tasks, but deeper assessments must consider conceptual plausibility, historiographic alignment, and the model's ability to raise meaningful questions. Building better datasets and benchmarks is thus both a methodological and epistemic commitment: it requires deciding what to formalize, which disagreements to preserve, and how to balance structure with openness. Adopting open science practices, such as transparent annotation protocols, permissive licensing, and shared code, can foster collective reflection and support more robust, sustainable infrastructures for computational HPSS.

## 6.4 Lesson 4: LLMs should enhance, not replace, HPSS methodologies

LLMs offer powerful tools for working with scientific texts, but must be integrated in ways that support, rather than displace, HPSS's interpretive and reflexive commitments. In domains focused on automation or efficiency, outputs are often judged by predictive accuracy alone. In HPSS, by contrast, the stakes lie in interpretation: claims must be situated, contestable, and accountable.

For structured tasks with limited ambiguity, like scientific entity recognition, LLMs may soon yield outputs reliable enough for integration with minimal oversight. Yet these systems are never neutral: their behavior reflects training data, prompt design, and modeling assumptions. As with human collaborators, their outputs must remain open to scrutiny, especially when they inform interpretive claims.

For open-ended tasks, like topic modeling or citation context analysis, interpretive ambiguity is inherent. Here, LLMs can surface patterns or plausible framings, but these remain just that: interpretations, requiring contextual judgment. Their value lies not in providing "right" answers, but in provoking informed judgment.

In this sense, LLMs may serve as limited co-reasoners. Philosophers of science, for instance, might use them to simulate opposing views or expose hidden assumptions. While these systems do not, and may never, "understand" in a human sense, they can enrich exploration by generating plausible responses and alternatives.

More broadly, LLMs offer a potential bridge between qualitative and quantitative approaches, enabling close engagement with large corpora in ways that increasingly approximate the conceptual depth of traditional qualitative analysis. While they may never reach the same level of contextual sensitivity or interpretive nuance, they can support forms of large-scale textual inquiry that remain grounded in interpretive reasoning.

Ultimately, LLMs can expand the HPSS methodological repertoire, enabling new kinds of comparison, scaling interpretive analysis, and supporting hypothesis generation. But their value hinges on how they are used: sometimes as semi-autonomous tools, sometimes as prompts for reflection. Like HPSS itself, they rest on the premise that meaning is contextual. LLMs can help make that premise computationally tractable, but not epistemically settled.

## 6.5 Beyond methods: Infrastructures, values, and responsibilities

While this paper has focused on the methodological and interpretive opportunities LLMs bring to HPSS, it is equally important to acknowledge the ethical and political dimensions of their development and use. These models are embedded in infrastructures driven by commercial interests, trained on data that reflect social and cultural biases, and powered by energy-intensive computation with significant environmental costs ([Bender et al., 2021](#); [Strubell et al., 2019](#)). Their growing integration into scholarly workflows via proprietary APIs, retrieval systems, and citation tools, raises concerns about platform dependency, infrastructural opacity, and the concentration of control over research infrastructure. HPSS scholars, in turn, have a responsibility to engage with LLMs critically, using them in ways that are ethically informed, environmentally conscious, and consistent with the field's interpretive and critical commitments.

Reflexive engagement with LLMs thus means looking beyond questions of utility or performance to consider who builds these models, under what institutional and material conditions, and to what ends. It also entails interrogating the assumptions embedded in the infrastructures through which they are deployed: assumptions about access, authority, authorship, and evaluation. As these systems increasingly influence how scholarly knowledge is produced and circulated, HPSS researchers face a dual responsibility: to attend to the values encoded in these tools, and to help define the norms, policies, and institutional structures that govern their development and use. In this light, adopting LLMs is not only a methodological choice but also an infrastructural and political act, calling for collective reflection and ethical accountability.

## 7. Conclusion

As we have shown, LLMs offer new possibilities for HPSS research, but they also raise new challenges. Their growing integration into scholarly workflows compels us to consider not just how they are used, but how they shape the practices and infrastructures through which knowledge is produced and evaluated.

Crucially, HPSS is well equipped to engage these challenges. The field has long examined how science is co-constituted by social, historical, and material conditions, and how knowledge is constructed, legitimized, and contested. The same analytical sensibilities and interpretive frameworks are essential for interrogating the assumptions embedded in LLMs and the systems that sustain them. HPSS offers a uniquely situated perspective to ask not only what LLMs do, but how and why, and whose interests they reflect.

Rather than treating LLMs as neutral tools or existential threats, we advocate a generative engagement: one that combines methodological experimentation with critical reflection, integrating computational approaches without abandoning interpretive depth. This dual orientation, both practical and reflexive, can guide how LLMs are used in HPSS and how they are understood more broadly.

LLMs are increasingly recognized as more than mere tools for processing text, though their exact nature remains open. Like other scientific models ([Morgan and Morrison, 1999](#)), they operate through abstraction, idealization, and representational design; they do not mirror language or meaning, but mediate it. Their internal structures encode assumptions about similarity, context, and relevance, influencing what can be noticed, retrieved, or inferred. In this sense, they are both objects of philosophical analysis and methodological provocations: What kind of models are they? Can their outputs be meaningfully described as interpretations? Do they understand and reason, or merely simulate surface forms? And how might engaging with these systems sharpen—or even revise—our concepts of interpretation, understanding, and reasoning in both human and machine-mediated contexts? These questions point not only to the limits of LLMs, but to the philosophical work they catalyze.

## Acknowledgments

We thank Gerd Graßhoff for numerous insightful discussions on the role of large language models in the history, philosophy, and sociology of science, as well as for his helpful comments on an early draft of this paper. We are also grateful to the participants of the interdisciplinary workshop *"Large Language Models for the History, Philosophy, and Sociology of Science"*, held at TU Berlin in April 2025. While the core research presented here predates and extends beyond the event, the workshop provided valuable perspectives that helped sharpen several of the questions addressed in this paper and offered important context for situating our analysis. This work was supported by the European Union under an ERC Consolidator Grant (Project No. 101044932, "Network Epistemology in Practice (NEPI)"). Views and opinions expressed are however those of the author only and do not necessarily reflect those of the European Union or the European Research Council. Neither the European Union nor the granting authority can be held responsible for them.

*Large Language Models for History, Philosophy, and Sociology of Science*Ji, Z., Wei, Q., & Xu, H. (2020). Bert-based ranking for biomedical entity normalization. *AMIA Summits on Translational Science Proceedings*, *2020*, 269.

Jiang, C., Xu, W., & Stevens, S. (2022). Edits: Understanding the Human Revision Process in Scientific Writing. In Y. Goldberg, Z. Kozareva, & Y. Zhang (Eds.), *Proceedings of the 2022 Conference on Empirical Methods in Natural Language Processing* (pp. 9420–9435). Association for Computational Linguistics.

Jiang, X., & Chen, J. (2023). Contextualised segment-wise citation function classification. *Scientometrics*, *128*(9), 5117–5158.

Just, J., Ströhle, T., Füller, J., & Hutter, K. (2024). AI-based novelty detection in crowdsourced idea spaces. *Innovation*, *26*(3), 359–386.

Kim, K., Kogler, D., & Maliphol, S. (2024). Identifying interdisciplinary emergence in the science of science: Combination of network analysis and BERTopic. *Humanities and Social Sciences Communications*, *11*(1), 1–15.

Kleymann, R., Niekler, A., & Burghardt, M. (2022). Conceptual Forays: A Corpus-based Study of "Theory" in Digital Humanities Journals. *Journal of Cultural Analytics*, *7*(4).

Knuuttila, T. (2011). Modelling and representing: An artefactual approach to model-based representation. *Studies in History and Philosophy of Science Part A*, *42*(2), 262–271.

Kunnath, S., Pride, D., & Knoth, P. (2022, November). *Dynamic Context Extraction for Citation Classification*. The 2nd Conference of the Asia-Pacific Chapter of the Association for Computational Linguistics and the 12th International Joint Conference on Natural Language Processing, Virtual.

Kunnath, S., Pride, D., & Knoth, P. (2023). Prompting Strategies for Citation Classification. *Proceedings of the 32nd ACM International Conference on Information and Knowledge Management*, 1127–1137.

Kutuzov, A., Velldal, E., & Øvrelid, L. (2022). Contextualized embeddings for semantic change detection: Lessons learned. *Northern European Journal of Language Technology*, *8*(1), Article 1.

Larooij, M., & Törnberg, P. (2025). *Do Large Language Models Solve the Problems of Agent-Based Modeling? A Critical Review of Generative Social Simulations*. arXiv:2504.03274.

Latour, B. (1987). *Science in action: How to follow scientists and engineers through society*. Harvard University Press.

Laubichler, M., Maienschein, J., & Renn, J. (2019). Computational History of Knowledge: Challenges and Opportunities. *Isis*, *110*(3), 502–512.

Lee, J., Yoon, W., Kim, S., et al. (2020). BioBERT: A pre-trained biomedical language representation model for biomedical text mining. *Bioinformatics*, *36*(4), 1234–1240.

Leto, A., Roy, S., Hoyle, A., et al. (2024). A First Step towards Measuring Interdisciplinary Engagement in Scientific Publications: A Case Study on NLP + CSS Research. In D. Card, A. Field, D. Hovy, & K. Keith (Eds.), *Proceedings of the Sixth Workshop on Natural Language Processing and Computational Social Science (NLP+CSS 2024)* (pp. 144–158). Association for Computational Linguistics.

Lewis, P., Perez, E., Piktus, A., et al. (2020). Retrieval-augmented generation for knowledge-intensive nlp tasks. *Advances in Neural Information Processing Systems*, *33*, 9459–9474.
24